\documentclass{article}

\usepackage[final]{nips_2018}

\usepackage[utf8]{inputenc} 
\usepackage[T1]{fontenc}    
\usepackage{hyperref}       
\usepackage{url}            
\usepackage{booktabs}       
\usepackage{amsfonts}       
\usepackage{nicefrac}       
\usepackage{microtype}      
\usepackage{microtype}
\usepackage{graphicx}
\usepackage{subcaption}
\usepackage{enumitem}
\usepackage{amsmath}
\usepackage{amssymb}
\usepackage{algorithm}
\usepackage{algpseudocode}

\def\Plus{\texttt{+}}
\def\Minus{\texttt{-}}
\newcommand{\pluseq}{\mathrel{+}=}

\title{Heterogeneous Bitwidth Binarization in \\Convolutional Neural Networks}

\author{
Josh Fromm \\
Department of Electrical Engineering \\
University of Washington \\
Seattle, WA 98195 \\
\texttt{jwfromm@uw.edu} \\
\And
Shwetak Patel \\
Department of Computer Science \\
University of Washington \\
Seattle, WA 98195 \\
\texttt{shwetak@cs.washington.edu} \\
\And
Matthai Philipose \\
Microsoft Research \\
Redmond, WA 98052 \\
\texttt{matthaip@microsoft.com} \\
}

\begin{document}

\maketitle

\begin{abstract}
Recent work has shown that fast, compact low-bitwidth neural networks can be
surprisingly accurate. These networks use {\em homogeneous binarization}:
all parameters in each layer or (more commonly) the whole model
have the same low bitwidth (e.g., 2 bits). However, modern hardware
allows efficient designs where each arithmetic instruction can have a
custom bitwidth, motivating {\em heterogeneous binarization}, where every
parameter in the network may have a different bitwidth. In this paper,
we show that it is feasible and useful to select bitwidths at the
parameter granularity during training. For instance a heterogeneously
quantized version of modern networks such as AlexNet and MobileNet,
with the right mix of 1-, 2- and 3-bit parameters that average
to just 1.4 bits can equal the accuracy of homogeneous 2-bit versions
of these networks. Further, we provide analyses to show that the
heterogeneously binarized systems yield FPGA- and ASIC-based
implementations that are correspondingly more efficient in both
circuit area and energy efficiency than their homogeneous counterparts.

\end{abstract}

\section{Introduction}
With Convolutional Neural Networks (CNNs) now outperforming humans in
vision classification tasks~\citep{szegedy2015going}, it is 
clear that CNNs will be a mainstay of AI
applications. However, CNNs are known to be
computationally demanding, and are most comfortably run on GPUs. For
execution in mobile and embedded settings, or when a given CNN is
evaluated many times, using a GPU may be too costly.
The search for inexpensive variants of CNNs has yielded techniques
such as hashing~\citep{chen2015compressing}, vector
quantization~\citep{gong2014compressing}, and
pruning~\citep{han2015deep}. One particularly promising track is
binarization ~\citep{courbariaux2015binaryconnect}, which
replaces 32-bit floating point values with single bits, either
\Plus 1 or \Minus 1, and (optionally) replaces floating point multiplies with
packed bitwise popcount-xnors~\cite{hubara2016quantized}.  Binarization can reduce
the size of models by up to 32$\times$, and reduce the number of
operations executed by up to 64$\times$. 

\par
It has not escaped hardware designers that the popcount-xnor operations
used in a binary network are especially well suited for
FPGAs or ASICs. Taking the xnor of two bits requires a single logic gate
compared to the hundreds required for even the most efficient floating point
multiplication units~\citep{ehliar2014area}. The drastically reduced area
requirements allows binary networks to be implemented with fully parallel
computations on even relatively inexpensive FPGAs~\citep{umuroglu2017finn}.
The level of parallelization afforded by these custom implementations allows them
to outperform GPU computation while expending a fraction of the power,
which offers a promising avenue of moving state of the art architectures
to embedded environments. We seek to improve the occupancy, power, and/or
accuracy of these solutions.

\par
Our approach is based on the simple observation that the power consumption, space needed,
and accuracy of binary models on FPGAs and custom hardware are proportional
to $mn$, where $m$ is the number of bits used to binarize input activations and $n$
is the number of bits used to binarize weights. Current binary algorithms restrict $m$ and $n$
to be integer values, in large part because efficient CPU implementations require parameters within
a layer to be the same bitwidth. However, hardware has no such requirements. Thus, we ask
whether bitwidths can be fractional. To address this question, we introduce
Heterogeneous Bitwidth Neural Networks (HBNNs), which allow \textit{each individual parameter
to have its own bitwidth}, giving a fractional average bitwidth to the model.

\par
Our main contributions are:  
\begin{enumerate}[label={(\arabic*)}]
\item We propose the problem of selecting the bitwidth of individual parameters
  during training such that the bitwidths average out to a specified value.
\item We show how to augment a state-of-the-art homogeneous binarization training scheme
  with a greedy bitwidth selection technique (which we call
  ``middle-out'') and a simple hyperparameter search to produce
  good heterogeneous binarizations efficiently.
\item
  We present a rigorous empirical evaluation (including on highly
  optimized modern networks such as Google's MobileNet) to show that
  heterogeneity yields equivalent accuracy at significantly lower
  average bitwidth.
\item
  Although implementing HBNNs efficiently on CPU/GPU may be difficult, we provide
  estimates based on recently proposed FPGA/ASIC implementations that
  HBNNs' lower average bitwidths can translate to significant
  reductions in circuit area and power. 

\end{enumerate}


\section{Homogeneous Network Binarization}
In this section we discuss existing techniques for binarization.
Table~\ref{related-work} summarizes their accuracy.\footnote{In line with 
  prior work, we use the AlexNet model trained on the ImageNet dataset as
  the baseline.}  

\par
When training a binary network, all techniques including ours maintain
weights in floating point format. During forward propagation, the
weights (and activations, if both weights and activations are to be
binarized) are passed through a {\em binarization function}
$\mathcal{B}$, which projects incoming values to a small, discrete
set. In backward  
propagation, a {\em custom gradient}, which updates the floating point
weights, is applied to the binarization layer. After training is
complete, the binarization function is 
applied one last time to the floating point weights to create a true
binary (or more generally, small, discrete) set of weights, which is
used for inference from then on.  
\par
Binarization was first introduced
by~\citet{courbariaux2015binaryconnect}. In this initial
investigation, dubbed BinaryConnect, 32-bit tensors $T$ were
converted to 1-bit variants $T^B$ using the stochastic equation 
\begin{equation}
  \mathcal{B}(T) \triangleq T^B =
  \begin{cases}
    $\Plus 1$ & \text{with probability $p = \sigma(T)$,} \\
    $\Minus 1$  & \text{with probability $ 1 - p$}
  \end{cases}
\end{equation}
where $\sigma$ is the hard sigmoid function defined by $\sigma(x) =$
max($0$, min($1$, $\frac{x + 1}{2}$)). For the custom gradient
function, BinaryConnect simply used $\frac{d{T^B}}{dT} = 1$.

Although BinaryConnect showed excellent results on relatively simple
datasets such as CIFAR-10 and MNIST, it performed poorly on
ImageNet, achieving only an 
accuracy of 27.9\%. ~\citet{courbariaux2016binarized} later improved
this model by simplifying the binarization by simply taking $T^B =
\text{sign}(T)$ and adding a gradient for this operation, namely the
{\em straight-through estimator}: 
\begin{equation}
\frac{d{T^B}}{dT} = 1_{|T| \leq 1}.
\end{equation}
The authors showed that the straight-through estimator allowed the binarization
of activations as well as weights without collapse of model performance. 
However, they did not attempt to train a model on ImageNet in this work.  
\par
\citet{rastegari2016xnor} made a slight modification to the simple pure 
single bit representation that showed improved results. Now taking a binarized approximation as
\begin{equation}
\label{straight-through}
	T^B = \alpha_i \text{sign}(T) \quad \text{with} \quad \alpha_i = \frac{1}{d} \sum_{j=1}^d |T_j|.
\end{equation}
This additional scalar term allows binarized values to better fit the
distribution of the incoming floating-point values, giving a higher
fidelity approximation for very little extra computation. The addition
of scalars and the straight-through estimator gradient allowed the
authors to achieve a Top-1 accuracy of 44.2\% on ImageNet. 
\par
\begin{figure}[t]
\begin{center}
\includegraphics[scale=0.22]{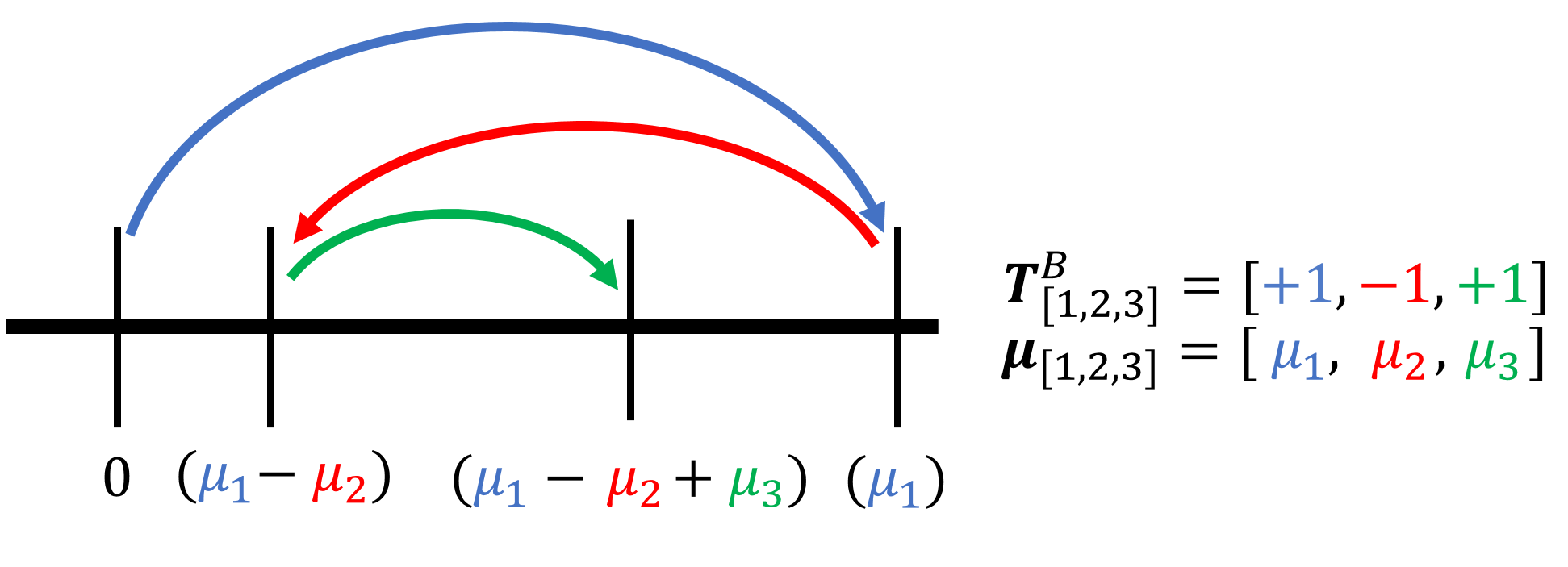}
\end{center}
\vspace{-0.15in}
\caption{Residual error binarization with $n=3$ bits. Computing each 
  bit takes a step from the position of the previous bit (see
  Equation~\ref{multibit_eq}).}
\vspace{-0.2in}
\label{binsteps}
\end{figure}

\citet{hubara2016quantized} and~\citet{zhou2016dorefa} found that
increasing the number of bits used to quantize the activations of the
network gave a considerable boost to the accuracy, achieving similar
Top-1 accuracy of 51.03\% and 50.7\% respectively. The precise
binarization function varied, but the typical approaches
include linearly or logarithmically placing the quantization points between 0 and 1,
 clamping values below a threshold distance from zero to
zero~\citep{li2016ternary}, and computing higher bits by measuring the
residual error from lower bits \citep{tang2017train}. All $n$-bit
binarization schemes require similar amounts of computation at
inference time, and have similar accuracy (see
Table~\ref{related-work}). In this work, we  extend
the {\em residual error binarization} function~\citet{tang2017train}
for binarizing to multiple ($n$) bits:
\begin{equation}
\label{multibit_eq}
\begin{gathered}
	T^B_{1} = \text{sign}(T),~\mu_1 = \text{mean}(|T|)\\
	E_{n} = T - \sum_{i=1}^n \mu_i \times T^B_{i} \\
	T^B_{n > 1} = \text{sign}(E_{n-1}),~\mu_{n > 1} = \text{mean}(|E_{n-1}|)\\
	T \approx \sum_{i = 1}^{n} \mu_i \times T^B_i
\end{gathered}
\end{equation}
where $T$ is the input tensor, $E_n$ is the
residual error up to bit $n$, $T^B_{n}$ is a tensor representing the $n^{\text{th}}$ bit
 of the approximation, and $\mu_n$ is a scaling factor for the
$n^{\text{th}}$ bit.  Note that the calculation of bit $n$ is
a recursive operation that relies on the values of all bits less than
$n$. Residual error binarization has each additional bit
take a step from the value of the previous bit. Figure~\ref{binsteps}
illustrates the process of binarizing a single value to 3 bits. Since
every binarized value is derived by taking $n$ steps, where each step
goes left or right, residual error binarization approximates inputs using
one of $2^n$ values.


\section{Heterogeneous Binarization}
\label{heterogeneousbinarization}
To date, there remains a considerable gap between the performance of
1-bit and 2-bit networks (compare rows 8 and 10 of
Table~\ref{related-work}). The highest full (i.e., where both weights
and activations are quantized) single-bit performer on AlexNet, 
Xnor-Net, remains roughly 7 percentage points less accurate (top 1) than the
2-bit variant, which is itself about 5.5 points less accurate than the
32-bit variant (row 25). When only weights are binarized, very recent
 results~\citep{Dong2017} similarly find that binarizing to 2 bits can
 yield nearly full accuracy (row 2), 
while the 1-bit equivalent lags by 4 points (row 1). The flip side to
using 2 bits for binarization is that the resulting models require
double the number of operations as the 1-bit variants at inference
time.

These observations naturally lead to the question, explored in this section, of whether it
is possible to attain accuracies closer to those of 2-bit models
while running at speeds closer to those of 1-bit variants. Of course,
it is also fundamentally interesting to understand whether it is
possible to match the accuracy of higher bitwidth models with those
that have lower (on average) bitwidth. Below, we discuss how
to extend residual 
error binarization to allow heterogeneous (effectively fractional)
bitwidths and present a method for distributing the
bits of a heterogeneous approximation. 

\subsection{Heterogeneous Residual Error Binarization via a Mask Tensor}
We modify Equation~\ref{multibit_eq}~, which binarizes to $n$ bits, to
instead binarize to a mixture of bitwidths by changing the third
line as follows:
\begin{equation}
\label{hetero_eq}
\begin{gathered}
	T^B_{n > 1} = \text{sign}(E_{n-1, j}),~\mu_{n > 1} = \text{mean}(|E_{n-1, j}|)\\~\text{with } j:M_j \geq n
\end{gathered}
\end{equation}
Note that the only addition is the {\em mask tensor} $M$, which is the same shape as
$T$, and specifies the number of bits $M_j$ that the $j^\text{th}$
entry of $T$ should be binarized to. In each round $n$ of the binarization
recurrence, we now only consider values that are not finished
binarizing, i.e, which have $M_j \geq n$. Unlike homogeneous
binarization, therefore, heterogeneous binarization generates
binarized values by taking {\em up to}, not necessarily exactly, $n$
steps. Thus, the number of distinct values representable is
$\sum_{i=1}^n 2^i = 2^{n+1} - 2$, which is roughly double that of the
homogeneous binarization.

In the homogeneous case, on
average, each step improves the accuracy of the approximation, but
there may be certain individual values that would benefit from not
taking a step, in Figure~\ref{binsteps} for example, it is possible
that ($\mu_1 - \mu_2$)  approximates the target value better
than ($\mu_1 - \mu_2 + \mu_3$). If values that benefit from not taking
a step can be targeted and assigned fewer bits, the overall
approximation accuracy will  improve despite there being a
lower average bitwidth.  

\subsection{Computing the Mask Tensor $M$}
\label{bitselectionmethods}
The question of how to distribute bits in a heterogeneous binary
tensor to achieve high representational power is equivalent to asking
how $M$ should be generated. When computing $M$, our goal is to take
an average bitwidth $B$ and determine both what fraction $P$  of $M$ should
be binarized to each bitwidth (e.g., $P =$ 5\% 3-bit, 10\% 2-bit and 85\%
1-bit for an average of $B=1.2$ bits), and how to distribute these bitwidths
across the individual entries in $M$. The full computation of $M$ is described
in Algorithm~\ref{M_gen}.
\par
We treat the distribution $P$ over
bitwidths as a model-wide hyperparameter. Since we only
search up to 3 bitwidths in practice, we perform a simple
grid sweep over the values of $P$. As we discuss
in Section~\ref{bitdistribution}, our discretization is relatively insensitive to these
hyperparameters, so a coarse sweep is adequate.
The results of the sweep are represented by the
function $DistFromAvg$ in Algorithm~\ref{M_gen}. 

\begin{figure*}[b!]
\centering
\begin{subfigure}{0.49\linewidth}
\centering
\includegraphics[width=\linewidth]{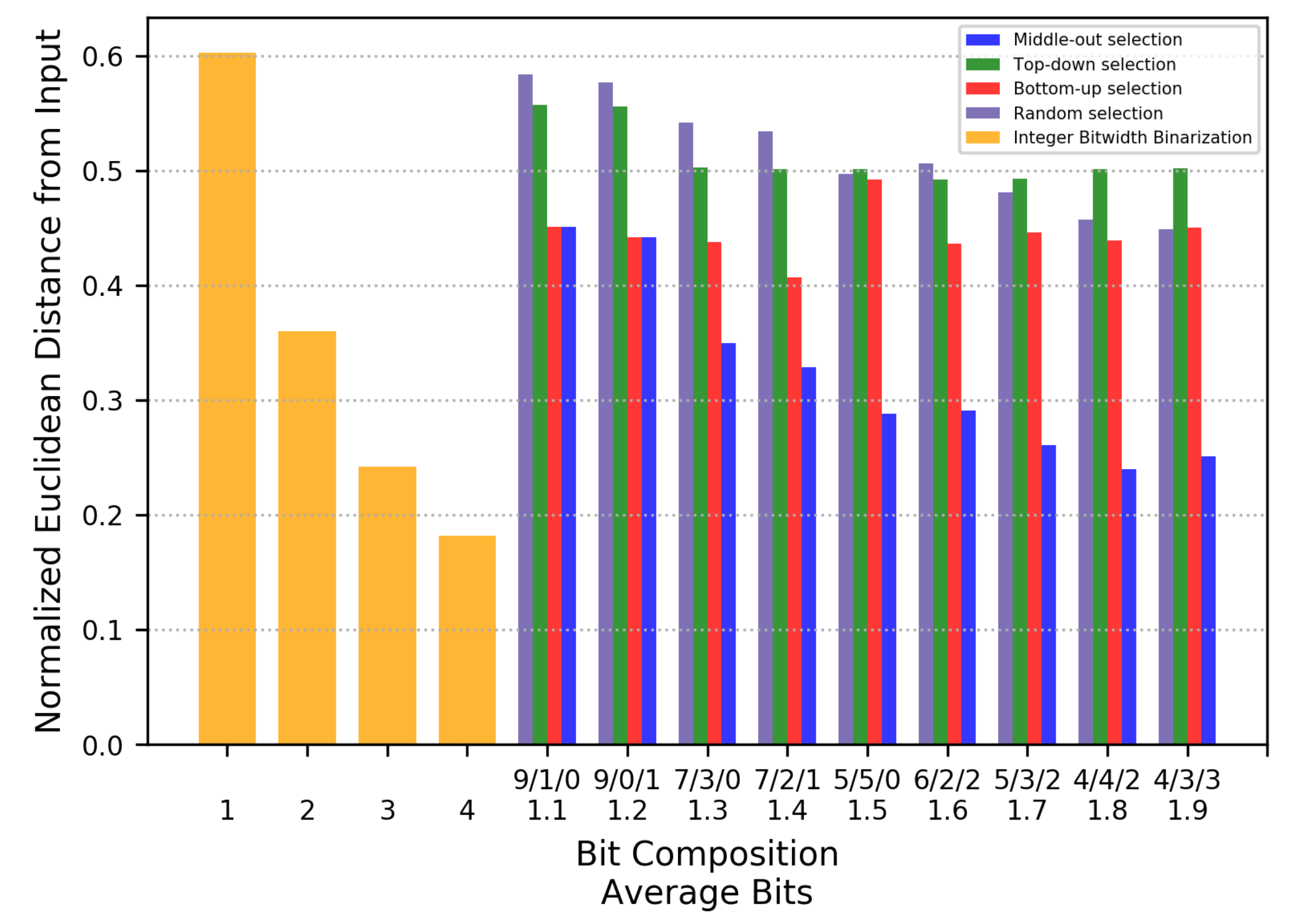}
\caption{Bit selection representational power.}
\label{euclid_distance}
\end{subfigure}
~
\begin{subfigure}{0.49\linewidth}
\centering
\includegraphics[width=\linewidth]{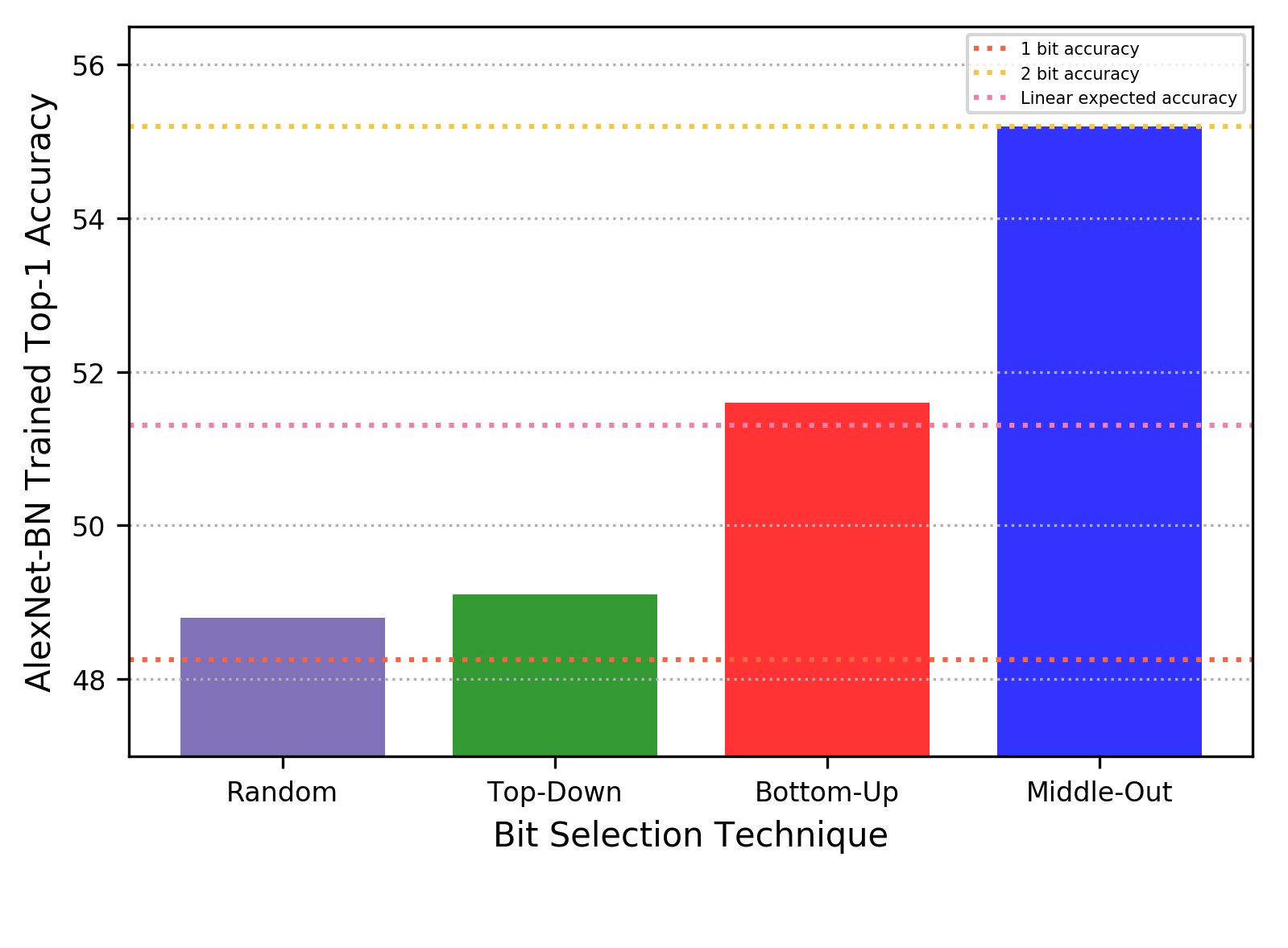}
\caption{1.4 bit HBNN AlexNet Accuracy. }
\label{alexnet_selection}
\end{subfigure}
\caption{Effectiveness of heterogeneous bit selection techniques (a)
  ability of different binarization schemes to approximate a large
  tensor of normally distributed random values. (b) accuracy of 1.4 bit heterogeneous
  binarized AlexNet-BN trained using each bit-selection technique.} 
\end{figure*}

Given $P$, we need to determine how to distribute the various bitwidths
using a value aware method: assigning low bitwidths to values that do not
need additional approximation and high bitwidths to those that do.
To this end, we propose several sorting heuristic methods: Top-Down (TD),
Middle-Out (MO), Bottom-Up (BU), and Random (R). These methods all attempt
to sort values of $T$ based on how many bits that value should be binarized
with. For example, Top-Down sorting assumes that larger values need fewer bits,
and so performs a standard descending sort. Similarly, Middle-Out sorting distributes
fewer bits to values closest to the mean of $T$, while Bottom-Up sorting assigns fewer
bits to smaller values. As a simple we control, we also consider Random sorting,
which assigns bits in a completely uninformed way. The definitions for the sorting
heuristics is given by Equation~\ref{selection_eqs}.
\begin{equation}
\label{selection_eqs}
\begin{split}
    &\mathrm{TD}(T) = \text{sort}(|T|, \text{descending})\\
    &\mathrm{MO}(T) = \text{sort}(|T| - \text{mean}(|T|), \text{ascending})\\
    &\mathrm{BU}(T) = \text{sort}(|T|, \text{ascending})\\
    &\mathrm{R}(T) = \text{a fixed uniformly random permutation of }T
\end{split}
\end{equation}

To evaluate the methods in Equation~\ref{selection_eqs}, we performed two experiments.
In the first, we create a large tensor of normally distributed values and binarize it with
a variety of bit distributions $P$ and each of the sorting heuristics using Algorithm~\ref{M_gen}. We then computed the Euclidean
distance between the binarized tensor and the original full precision tensor. A lower normalized distance
suggests a more powerful sorting heuristic. The results of this experiment are shown in Figure~\ref{euclid_distance}, and show
that Middle-Out sorting outperforms other heuristics by a significant margin. Notably, the results suggest that using Middle-Out sorting
can produce approximations with fewer than 2-bits that are comparably accurate to 3-bit integer binarization.

To confirm these results translate to accuracy in binarized convolutional networks, we consider 1.4 bit binarized AlexNet, with bit
distribution $P$ set to 70\% 1-bit, 20\% 2-bit, and 10\% 3-bit, an average of 1.4 bits. The specifics of the model and training procedure are the same as those
described in Section~\ref{implementation_details}. We train this model with each of the sorting
heuristics and compare the final accuracy to gauge the representational strength of each heuristic. The results are shown in Figure~\ref{alexnet_selection}.
As expected, Middle-Out sorting performs significantly better than other heuristics and yields an accuracy comparable to 2-bit integer binarization despite
using on average 1.4 bits.

The intuition behind the exceptional performance of Middle-Out is based on Figure~\ref{binsteps}~. We can see that the values that 
are most likely to be accurate without additional bits are those that are closest to the average $\mu_n$
for each step $n$. By assigning low bitwidths to the most average values, we can not just minimize
losses, but in some cases provide a better approximation using fewer average steps. In proceeding sections, all training and evaluation
is performed with Middle-Out as the sorting heuristic in Algorithm~\ref{M_gen}.

\begin{algorithm}[t!]
  \caption{Generation of bit map $M$.}
  \textbf{Input:} A tensor $T$ of size $N$ and an average bitwidth $B$.\\
  \textbf{Output:} A bit map $M$ that can be used in Equation~\ref{hetero_eq} to heterogeneously binarize $T$. 
  \begin{algorithmic}[1]
	\State $R$ = $T$ \Comment{Initialize $R$, which contains values that have not yet been assigned a bitwidth}
  	
  	\State $x=0$
    \State $P = \text{DistFromAvg}(B)$ \Comment Generate distribution of bits to fit average.
  	\For{($b$, $p_b$) in $P$} \Comment{$b$ is a bitwidth and $p_b$ is the percentage of $T$ to binarize to width $b$.}
		\State $S = \text{SortHeuristic}(R)$ \Comment{Sort indices of
                  remaining values by suitability for $b$-bit  binarization.}
  		\State $M[S[x : x+ p_bN]] = b$
		\State $R = R \setminus R[S[x : x+ p_bN]]$ \Comment Do not consider these indices in next step. 
  		\State $x \pluseq p_bN$
    \EndFor
  \end{algorithmic}
  \label{M_gen}
\end{algorithm}

\section{Experiments}
\label{experiments}
To evaluate HBNNs we wished to answer the following three questions:
\begin{enumerate}[label={(\arabic*)}]
\item How does accuracy scale with an uninformed bit distribution? 
\item How well do HBNNs perform on a challenging dataset compared to the state of the art?
\item Can the benefits of HBNNs be transferred to other architectures?
\end{enumerate}
In this section we address each of these questions.
\subsection{Implementation Details} 
\label{implementation_details}
AlexNet with batch-normalization (AlexNet-BN) is the standard model
used in binarization work due to its longevity and the general
acceptance that improvements made to accuracy transfer well to more
modern architectures. Batch normalization layers are applied to the
output of each convolution block, but the model is otherwise identical
to the original AlexNet model proposed
by~\citet{krizhevsky2012imagenet}. Besides it's benefits in improving
convergence, \citet{rastegari2016xnor} found that batch-normalization is especially important for binary
networks because of the need to equally distribute values around
zero. We additionally insert binarization functions within the
convolutional layers of the network when binarizing weights and at the
input of convolutional layers when binarizing inputs. We keep a
floating point copy of the weights that is updated during
back-propagation, and binarized during forward propagation as is
standard for binary network training. We use the straight-through
estimator for gradients.

When binarizing the weights
of the network's output layer, we add a single parameter scaling layer
that helps reduce the numerically large outputs of a binary layer to a
size more amenable to softmax, as suggested
by~\citet{tang2017train}. We train all models using an SGD solver with
learning rate 0.01, momentum 0.9, and weight decay 1e-4 and randomly
initialized weights for 90 epochs on PyTorch. 

\begin{figure*}[b!]
\centering
\begin{subfigure}{0.49\linewidth}
\centering
\includegraphics[width=\linewidth]{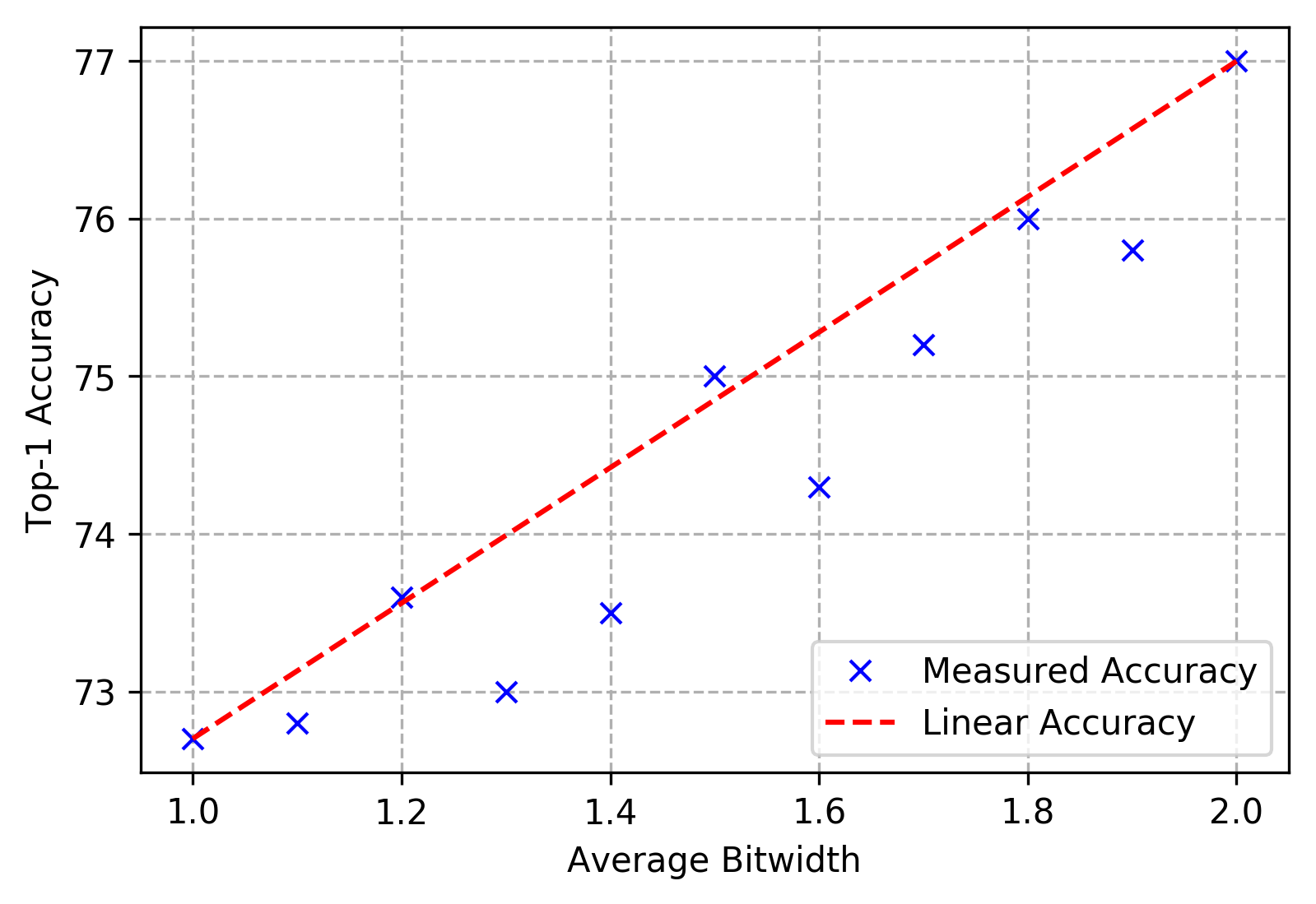}
\caption{CIFAR-10 uninformed bit selection.}
\label{cifar10}
\end{subfigure}
~
\begin{subfigure}{0.49\linewidth}
\centering
\includegraphics[width=\linewidth]{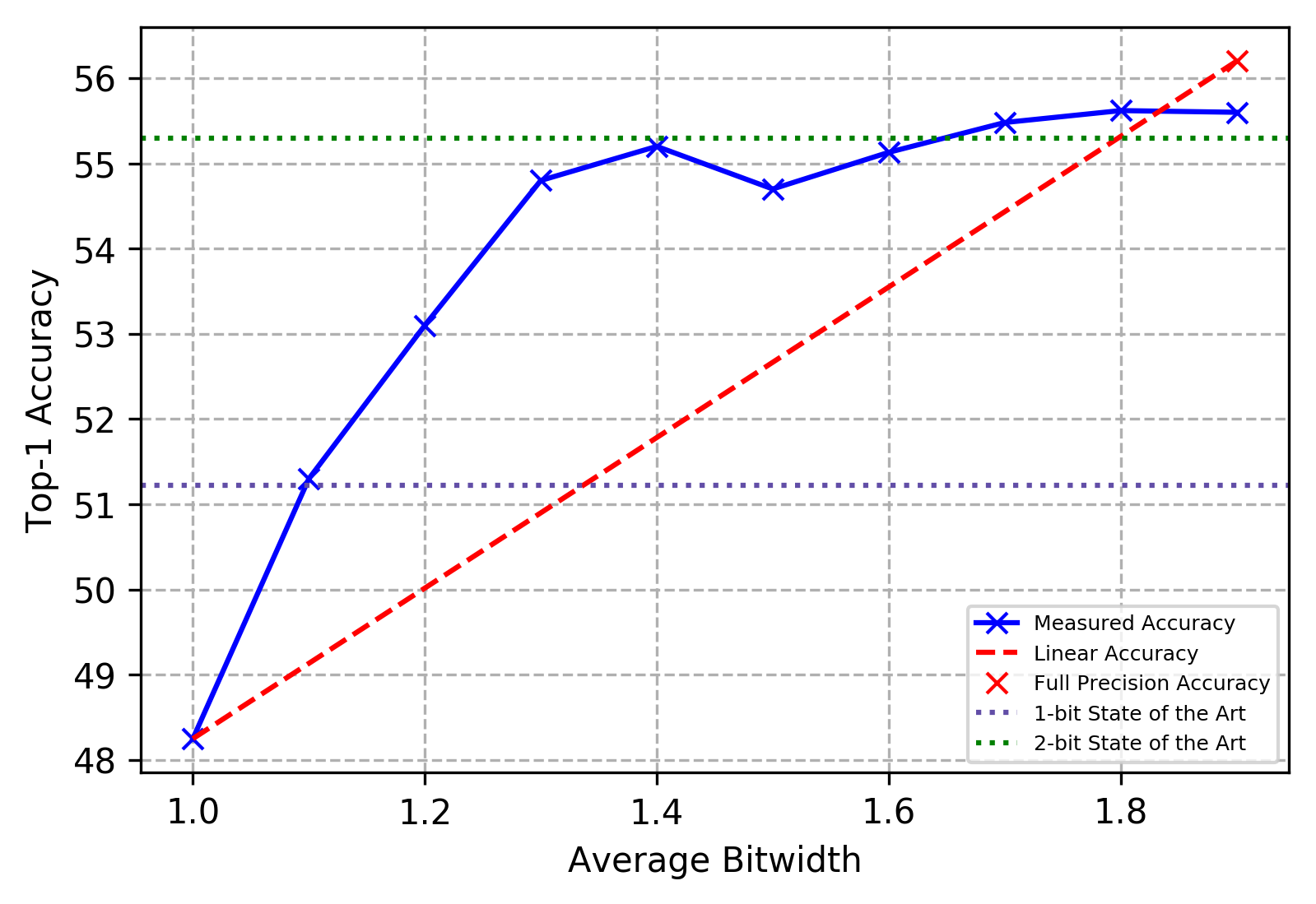}
\caption{HBNN AlexNet with Middle-Out bit selection. }
\label{binw_acc}
\end{subfigure}
\caption{Accuracy results of trained HBNN models. (a) Sweep of heterogenous bitwidths 
on a deliberately simplified four layer convolutional model for CIFAR-10. 
(b) Accuracy of heterogeneous bitwidth AlexNet-BN models. Bits are distributed using the 
Middle-Out selection algorithm.}
\end{figure*}


\subsection{Layer-level Heterogeneity}
As a baseline, we test a ``poor man's'' approach to HBNNs, where we
 fix up front the number of bits each layer is allowed, require all
 values in a layer to have its associated bitwidth, and
then train as with conventional homogeneous binarization. We consider
10 mixes of 1, 2 and 3-bit layers so as to sweep average bitwidths
between 1 and 2. We trained as described in
Section~\ref{implementation_details}. For this experiment, we used the
CIFAR-10 dataset with a deliberately hobbled (4-layer fully
convolutional) model  with a maximum accuracy of roughly 78\% as the
baseline 32-bit variant. We chose CIFAR-10  to allow quick
experimentation. We chose not to use a large model for CIFAR-10,
because for large models it is known that even 1-bit models have
32-bit-level accuracy~\citet{courbariaux2016binarized}.

Figure~\ref{cifar10} shows the results. Essentially, accuracy 
increases roughly linearly with average bitwidth. Although such linear
scaling of accuracy with bitwidth is itself potentially useful (since
it allows finer grain tuning on FPGAs), we are hoping for even better
scaling with the ``data-aware'' bitwidth selection provided by HBNNs.
\subsection{Bit Distribution Generation}
\label{bitdistribution}
As described in~\ref{bitselectionmethods}, one of the considerations when
using HBNNs is how to take a desired average bitwidth and produce a matching
distribution of bits. For example, using 70\% 1-bit, 20\% 2-bit and 10\% 3-bit
values gives an average of 1.4 bits, but so too does 80\% 1-bit and 20\% 3-bit values.
We suspected that the choice of this distribution would have a significant impact
on the accuracy of trained HBNNs, and performed a hyperparameter sweep by varying
$DistFromAvg$ in Algorithm~\ref{M_gen} when training AlexNet on ImageNet as described
in the following sections. However, much to our surprise, \textbf{models trained
with the same average bitwidth achieved nearly identical accuracies regardless of
distribution}. For example, the two 1.4-bit distributions given above yield accuracies of
49.4\% and 49.3\% respectively. This suggests that choice of $DistFromAvg$ is actually unimportant,
which is quite convenient as it simplifies training of HBNNs considerably.

\subsection{AlexNet: Binarized Weights and Non-Binarized Activations}
\label{subsec_ex_weights}
\begin{table*}[t]
\small
\caption{Accuracy of related binarization work and our results}
\label{related-work}
\begin{center}
\begin{tabular}{llllll}
\toprule
&\multicolumn{1}{l}{Model} 
&\multicolumn{1}{l}{Name} 
&\multicolumn{1}{c}{Binarization (Inputs / Weights)} 
&\multicolumn{1}{c}{Top-1} 
&\multicolumn{1}{c}{Top-5}\\
\midrule
\multicolumn{6}{c}{Binarized weights with floating point activations} \\
1 & AlexNet & SQ-BWN~\citep{Dong2017} & full precision / 1-bit & 51.2\% & 75.1\%\\
2 & AlexNet & SQ-TWN~\citep{Dong2017} & full precision / 2-bit & 55.3\% & 78.6\%\\
3 & AlexNet & TWN (our implementation) & full precision / 1-bit & 48.3\% & 71.4\%\\
4 & AlexNet & TWN  & full precision / 2-bit & 54.2\% & 77.9\%\\
5 & AlexNet & HBNN (our results) & full precision / 1.4-bit & 55.2\% & 78.4\%\\
6 & MobileNet & HBNN & full precision / 1.4-bit & 65.1\% & 87.2\%\\
\midrule
\multicolumn{6}{c}{Binarized weights and activations excluding input and output layers} \\
7 & AlexNet & BNN~\citep{courbariaux2015binaryconnect} & 1-bit / 1-bit & 27.9\% & 50.4\%\\
8 & AlexNet & Xnor-Net~\citep{rastegari2016xnor} & 1-bit / 1-bit & 44.2\% & 69.2\%\\
9 & AlexNet & DoReFaNet~\citep{zhou2016dorefa} & 2-bit / 1-bit & 50.7\% & 72.6\%\\
10 & AlexNet & QNN~\citep{hubara2016quantized} & 2-bit / 1-bit & 51.0\% & 73.7\%\\
11 & AlexNet & our implementation & 2-bit / 2-bit & 52.2\% & 74.5\%\\
12 & AlexNet & our implementation & 3-bit / 3-bit & 54.2\% & 78.1\%\\
13 & AlexNet & HBNN & 1.4-bit / 1.4-bit & 53.2\% & 77.1\%\\
14 & AlexNet & HBNN & 1-bit / 1.4-bit & 49.4\% & 72.1\%\\
15 & AlexNet &HBNN & 1.4-bit / 1-bit & 51.5\% & 74.2\%\\
16 & AlexNet &HBNN & 2-bit / 1.4-bit & 52.0\% & 74.5\%\\
17 & MobileNet & our implementation & 1-bit / 1-bit & 52.9\% & 75.1\%\\
18 & MobileNet & our implementation & 2-bit / 1-bit & 61.3\% & 80.1\%\\
19 & MobileNet & our implementation & 2-bit / 2-bit & 63.0\% & 81.8\%\\
20 & MobileNet & our implementation & 3-bit / 3-bit & 65.9\% & 86.7\%\\
21 & MobileNet & HBNN & 1-bit / 1.4-bit & 60.1\% & 78.7\%\\
22 & MobileNet & HBNN & 1.4-bit / 1-bit & 62.0\% & 81.3\%\\
23 & MobileNet & HBNN & 1.4-bit / 1.4-bit & 64.7\% & 84.9\%\\
24 & MobileNet & HBNN & 2-bit / 1.4-bit & 63.6\% & 82.2\%\\
\midrule
\multicolumn{6}{c}{Unbinarized (our implementation)}\\
25 & AlexNet & ~\citep{krizhevsky2012imagenet} & full precision / full precision & 56.5\% & 80.1\%\\
26 & MobileNet & ~\citep{howard2017mobilenets} & full precision / full precision & 68.8\% & 89.0\%\\
\bottomrule
\end{tabular}
\end{center}
\vspace{-0.2in}
\end{table*}
Recently,~\citet{Dong2017} were able to binarize the weights of an
AlexNet-BN model to 2 bits and achieve nearly full precision accuracy
(row 2 of Table~\ref{related-work}). We consider this to be the state
of the art in weight binarization since the model achieves excellent
accuracy despite all layer weights being binarized, including the
input and output layers which have traditionally been difficult to
approximate. We perform a sweep of AlexNet-BN models binarized with
fractional bitwidths using middle-out selection with the goal of
achieving comparable accuracy using fewer than two bits.

The results
of this sweep are shown in Figure~\ref{binw_acc}. We were able to
achieve nearly identical top-1 accuracy to the best full 2 bit results
(55.3\%) with an average of only 1.4 bits (55.2\%). As we had hoped, we also found that the accuracy 
scales in a {\em super-linear} manner with respect to bitwidth when
using middle-out 
bit selection. Specifically, the model accuracy increases extremely
quickly from 1 bit to 1.3 bits before slowly approaching the full
precision accuracy. 


\subsection{AlexNet: Binarized Weights and Activations}
In order to realize the speed-up benefits of binarization (on CPU or
FPGA) in practice, it is necessary to binarize both inputs the weights, which
allows floating point multiplies to be replaced with packed bitwise
logical operations. The number of operations in a binary network is 
reduced by a factor of $\frac{64}{mn}$ where $m$ is the number of bits
used to binarize inputs and $n$ is the number of bits to binarize
weights. Thus, there is significant motivation to keep the bitwidth of
both inputs and weights as low as possible without losing too much
accuracy. When binarizing inputs, the input and output layers are
typically not binarized as the effects on the accuracy are much larger
than other layers. We perform another sweep on AlexNet-BN with all
layers but the input and output fully binarized and compare the
accuracy of HBNNs to several recent results. Row 8 of
Table~\ref{related-work} is the top previously reported accuracy (44.2\%)
for single bit input and weight binarization, while row 10 (51\%) is the top
accuracy for 2-bit inputs and 1-bit weights.

\par
Table~\ref{related-work} (rows 13 to 16) reports a selection of
results from this search. Using 1.4 bits to binarize 
inputs and weights ($mn = 1.4\times1.4 = 1.96$)  gives a very high
accuracy (53.2\% top-1) while having the same number of total
operations $mn$ as a network, such as the one from row 10,
binarized with 2 bit activations and 1 bit weights. We have similarly
good results when leaving the input binarization bitwidth an
integer. Using 1 bit inputs and 1.4 bit weights, we reach 49.4\% top-1
accuracy which is a large improvement over~\citet{rastegari2016xnor}
at a small cost. We found that using more than 1.4 average bits had
very little impact on the overall accuracy. Binarizing inputs to 1.4
bits and weights to 1 bit (row 15) similarly
outperforms~\citet{hubara2016quantized} (row 10). 

\subsection{MobileNet Evaluation}
\label{mobileneteval}
Although AlexNet serves as an essential measure to compare to previous and related work,
it is important to confirm that the benefits of heterogeneous binarization is model independent.
To this end, we perform a similar sweep of binarization parameters on MobileNet, a state of
the art architecture that has unusually high accuracy for its low number of parameters~\citep{howard2017mobilenets}.
MobileNet is made up of separable convolutions instead of the typical dense convolutions of AlexNet. Each separable
convolution is composed of an initial spatial convolution followed by a depth-wise convolution. Because the vast bulk
of computation time is spent in the depth-wise convolution, we binarize only its weights, leaving the spatial weights floating point.
We binarize the depth wise weights of each MobileNet layer in a similar fashion as in section~\ref{subsec_ex_weights} and achieve
a Top-1 accuracy of 65.1\% (row 6). This is only a few percent below our unbinarized implementation (row 26), which is
an excellent result for the significant reduction in model size. 

\par
We additionally perform a sweep of many different binarization
bitwidths for both the depth-wise weights and input activations of MobileNet, with results shown in rows 17-24 of Table~\ref{related-work}. Just as in the AlexNet
case, we find that MobileNet with an average of 1.4 bits (rows 21 and 22) achieves over 10\% higher accuracy than 1-bit
binarization (row 17). We similarly observe that 1.4-bit binarization outperforms 2-bit binarization in each permutation of bitwidths.
The excellent performance of HBNN MobileNet confirms that heterogeneous binarization is fundamentally valuable, and
we can safely infer that it is applicable to many other network architectures as well.


\section{Hardware Implementability}
\label{fpga_section}
\begin{table*}[t]
\small
\caption{Hardware Implementation Metrics}
\label{fpga_metrics}
\begin{center}
\begin{tabular}{lllllllll}
\toprule
& \multicolumn{1}{l}{Platform} 
& \multicolumn{1}{l}{Model} 
& \multicolumn{1}{l}{Unfolding}
& \multicolumn{1}{l}{Bits} 
& \multicolumn{1}{l}{Occupancy} 
& \multicolumn{1}{l}{kFPS} 
& \multicolumn{1}{l}{$\text{P}_\text{chip}$ (W)}
& \multicolumn{1}{l}{Top-1}\\
\midrule
\multicolumn{9}{c}{CIFAR-10 Baseline Implementations} \\
1& ZC706 & VGG-8 & 1$\times$ & 1 & 21.2\% & 21.9 & 3.6 & 80.90\%\\
2& ZC706 & VGG-8 & 4$\times$ & 1 & 84.8\% & 87.6 & 14.4 & 80.90\%\\
3& ASIC & VGG-8 & - & 2 & 6.06 mm$^2$ & 3.4 & 0.38 & 87.89\%\\
\midrule
\multicolumn{9}{c}{CIFAR-10 HBNN Customization} \\
4 & ZC706 & VGG-8 & 1$\times$ & 1.2 & 25.4\% & 18.25 & 4.3 & 85.8\%\\
5 & ZC706 & VGG-8 & 1$\times$ & 1.4 & 29.7\% & 15.6 & 5.0 & 89.4\%\\
6 & ZC706 & VGG-8 & 4$\times$ & 1.2 & 100\% & 73.0 & 17.0 & 85.8\%\\
7 & ASIC & VGG-8 & - & 1.2 & 2.18 mm$^2$ & 3.4 & 0.14 & 85.8\%\\
8 & ASIC & VGG-8 & - & 1.4 & 2.96 mm$^2$ & 3.4 & 0.18 & 89.4\%\\
\midrule
\multicolumn{9}{c}{Extrapolation to MobileNet with ImageNet Data} \\
9 & ZC706 & MobileNet & 1$\times$ & 1 & 20.0\% & 0.45 & 3.4 & 52.9\%\\
10 & ZC706 & MobileNet & 1$\times$ & 2 & 40.0\% & 0.23 & 6.8 & 63.0\%\\
11 & ZC706 & MobileNet & 1$\times$ & 1.4 & 28.0\% & 0.32 & 4.76 & 64.7\%\\
12 & ASIC & MobileNet & - & 2 & 297 mm$^2$ & 3.4 & 18.62 & 63.0\%\\
13 & ASIC & MobileNet & - & 1.4 & 145.5 mm$^2$ & 3.4 & 9.1 & 64.7\%\\
\bottomrule
\end{tabular}
\end{center}
\vspace{-0.2in}
\end{table*}

Our experiments demonstrate that HBNNs have significant advantages
compared to integer bitwidth approximations. However, with these representational
benefits come added complexity in implementation. Binarization typically provides a 
significant speed up by packing bits into 64-bit integers, allowing a CPU or GPU to
perform a single xnor operation in lieu of 64 floating-point multiplications. However,
Heterogeneous tensors are essentially composed of sparse arrays of bits. Array sparsity
makes packing bits inefficient, nullifying much of the speed benefits one would expect
from having fewer average bits. The necessity of bit packing exists because CPUs and
GPUs are designed to operate on groups of bits rather than individual bits. However,
programmable or custom hardware such as FPGAs and ASICs have no such restriction.
In hardware, each parameter can have its own set of $n$ xnor-popcount units, where $n$ is the bitwidth
of that particular parameter. In FPGAs and ASICs, the total number of computational units
in a network has a significant impact on the power consumption and speed of inference. Thus,
the benefits of HBNNs, higher accuracy with fewer computational units, are fully realizable.

\par
There have been several recent binary convolutional neural network implementations on FGPAs
and ASICs that provide a baseline we can use to estimate the performance of HBNNs on
ZC706 FPGA platforms~\citep{umuroglu2017finn} and on ASIC hardware~\citep{alemdar2017ternary}. The results of these
implementations are summarized in rows 1-3 of Table~\ref{fpga_metrics}. Here, unfolding refers
to the number of computational units placed for each parameter, by having multiple copies of a 
parameter, throughput can be increased through improved parallelization. Bits refers to the level
of binarization of both the input activations and weights of the network. Occupancy is the number
of LUTs required to implement the network divided by the total number of LUTs available for an FPGA,
or the chip dimensions for an ASIC. Rows 4-12 of Table~\ref{fpga_metrics} show the metrics of HBNN
versions of the baseline models. Some salient points that can be drawn from the table include:
\begin{itemize}
\item Comparing lines 1, 4, and 5 show that on FPGA, fractional binarization offers fine-grained tuning of the
performance-accuracy trade-off. Notably, a significant accuracy boost is obtainable for only slightly higher occupancy
and power consumption.
\item Rows 2 and 6 both show the effect of unrolling. Notably,
with 1.2 average bits, there is no remaining space on the ZC706. This means that using a full 2 bits, a designer would have to use
a lower unrolling factor. In many cases, it may be ideal to adjust average bitwidth to reach maximum occupancy, giving the highest possible
accuracy without sacrificing throughput.
\item Rows 3, 7, and 8 show that in ASIC, the size and power consumption of a chip can be drastically reduced without impacting accuracy at all.
\item Rows 9-13 demonstrate the benefits of fractional binarization are not restriced to CIFAR, and extend to MobileNet in a similar way.
The customization options and in many cases direct performance boosts offered by HBNNs are valuable regardless of model architecture.
\end{itemize}


\section{Conclusion}
In this paper, we present Heterogeneous Bitwidth Neural Networks (HBNNs),
 a new type of binary network that is not restricted to integer bitwidths. Allowing
 effectively fractional bitwidths in networks gives a vastly improved ability to tune the 
trade-offs between accuracy, compression, and speed that come with binarization. 
We introduce middle-out bit selection as the top performing 
technique for determining where to place bits in a heterogeneous bitwidth tensor. 
On the ImageNet dataset with AlexNet and MobileNet models, we 
perform extensive experiments to validate the effectiveness of HBNNs compared to the 
state of the art and full precision accuracy. The results of these experiments are highly 
compelling, with HBNNs matching or outperforming competing binarization techniques 
while using fewer average bits.

\clearpage

\bibliography{HBNN_NIPS}
\bibliographystyle{icml2018}

\end{document}